\documentclass[letterpaper]{article} 
\usepackage{aaai24}  
\usepackage{times}  
\usepackage{helvet}  
\usepackage{courier}  
\usepackage[hyphens]{url}  
\usepackage{graphicx} 
\usepackage{natbib}  
\usepackage{caption} 
\usepackage{algorithm}
\usepackage{algorithmic}
\usepackage{booktabs}
\usepackage{multirow}
\usepackage{epstopdf}
\usepackage{subfigure}

\usepackage{newfloat}
\usepackage{listings}
\DeclareCaptionStyle{ruled}{labelfont=normalfont,labelsep=colon,strut=off} 
\floatstyle{ruled}
\newfloat{listing}{tb}{lst}{}
\floatname{listing}{Listing}
\title{BCLNet: Bilateral Consensus Learning for Two-View Correspondence Pruning}
\author {
    Xiangyang Miao\textsuperscript{\rm 1,2},
    Guobao Xiao\textsuperscript{\rm 1}\thanks{Corresponding author},
    Shiping Wang\textsuperscript{\rm 2},
    Jun Yu\textsuperscript{\rm 3}
}
\affiliations {
    \textsuperscript{\rm 1}School of Electronics and Information Engineering, Tongji University,\\
    \textsuperscript{\rm 2}College of Computer and Data Science, Fuzhou University, \\
    \textsuperscript{\rm 3}School of Computer Science and Technology, Hangzhou Dianzi University\\
    mxyttkx@163.com,
    x-gb@163.com,
    shipingwangphd@163.com,
    yujun@hdu.edu.cn
}

\usepackage{bibentry}

\begin{document}

\maketitle

\begin{abstract}
Correspondence pruning aims to establish reliable correspondences between two related images and recover relative camera motion. Existing approaches often employ a progressive strategy to handle the local and global contexts, with a prominent emphasis on transitioning from local to global, resulting in the neglect of interactions between different contexts. To tackle this issue, we propose a parallel context learning strategy that involves acquiring bilateral consensus for the two-view correspondence pruning task. In our approach, we design a distinctive self-attention block to capture global context and parallel process it with the established local context learning module, which enables us to simultaneously capture both local and global consensuses. By combining these local and global consensuses, we derive the required bilateral consensus. We also design a recalibration block, reducing the influence of erroneous consensus information and enhancing the robustness of the model. The culmination of our efforts is the Bilateral Consensus Learning Network (BCLNet), which efficiently estimates camera pose and identifies inliers (true correspondences). Extensive experiments results demonstrate that our network not only surpasses state-of-the-art methods on benchmark datasets but also showcases robust generalization abilities across various feature extraction techniques. Noteworthily, BCLNet obtains 3.98\% mAP5$^{\circ}$ gains over the second best method on unknown outdoor dataset, and obviously accelerates model training speed. The source code will be available at: https://github.com/guobaoxiao/BCLNet.
\end{abstract}

\section{Introduction}

\begin{figure}[t]
\centering
\centerline{\includegraphics[width=0.48\textwidth]{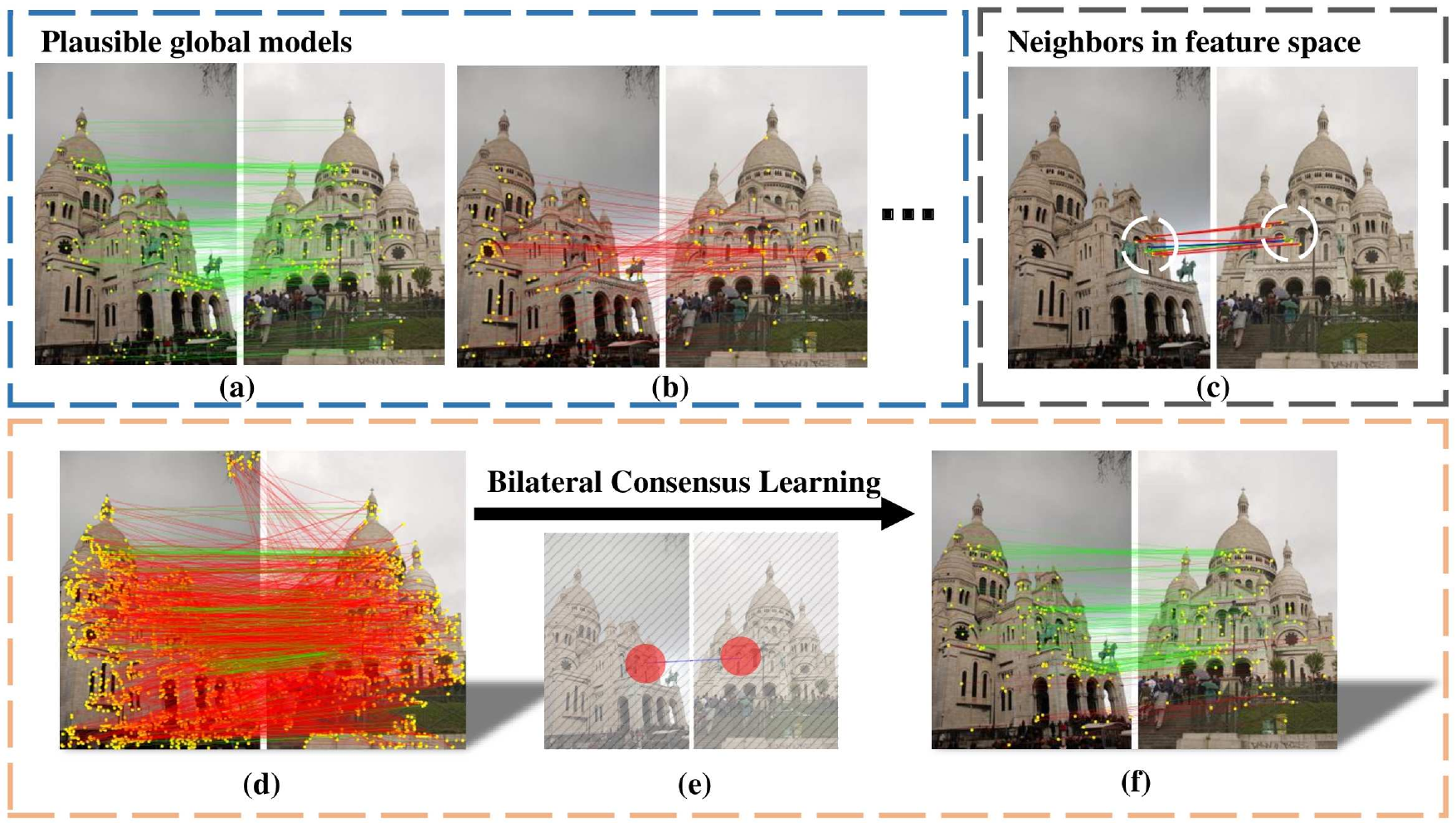}}
\caption{Bilateral consensus acquisition process. Both local and global contexts are inevitably affected by outliers, neglecting the interaction between them tends to exacerbate the propagation of erroneous information. There may be multiple models in the network that satisfy global constraints (a) and (b). Neighbors based on k-nearest neighbor search also contain many outliers (c). Given a set of putative correspondences (d), we adopt existing blocks and the designed BCMA block (e) to extract local and global consensuses, respectively. Subsequently, we facilitate their interaction to achieve bilateral consensus, which ultimately generates the network's prediction (f). The red lines represent outliers, green lines represent inliers, and blue represents selected correspondences.}
\label{fig:Introduce}
\end{figure}

Accurate two-view correspondences play a pivotal role in various computer vision applications, including stereo matching~\cite{Yao2021}, simultaneous localization and mapping (SLAM)~\cite{Kazerouni2022}, and  structure from motion~\cite{Xiao2021}, $etc$. Standard sparse matching pipeline relies on off-the-shelf methods (SIFT~\cite{Lowe2004} or Superpoint~\cite{DeTone2018}) to establish initial correspondences. However, these initial correspondences often harbor a considerable number of outliers (false correspondences) due to the intricate image variations~\cite{Jin2021,Ma2021}, such as illumination changes, repeated textures, image blurs and occlusions. Therefore, to alleviate the repercussions on downstream tasks, the incorporation of a correspondence pruning algorithm emerges as an indispensable stride.

Existing correspondence pruning algorithms can currently be categorized into two classes: traditional and learning-based algorithms. Among traditional algorithms, RANSAC~\cite{Fischler1981} and its diverse adaptations~\cite{Chum2005,Barath2019,Torr2000} are the most popular ones, which primarily rely on a generate-and-verify strategy. Although they demonstrate promising results on specific tasks, their theoretical execution time tends to experience exponential growth as the proportion of outliers increases~\cite{Zheng2022}.

Fortunately, the advancement of deep learning offers a novel solution for the correspondence pruning task. Following the pioneering work LFGC~\cite{Yi2018}, learning-based works mostly partition correspondence pruning into a correspondence classification task and a camera pose estimation task. Then, CLNet~\cite{Zhao2021} initially introduces a progressive local-to-global learning strategy and it has gradually become the default standard for subsequent researches. The most recent work, ConvMatch~\cite{Zhang2023}, firstly adopts convolutions to solve this problem by mapping correspondences onto grids, but it also abides by the progressive local-to-global strategy. It's worth mentioning that while this progressive strategy yields favorable results, a crucial aspect has frequently been neglected--the interaction between local and global contexts. As depicted in Figure~\ref{fig:Introduce} (a)-(c), challenging scenarios reveal that both local and global contexts are susceptible to significant outlier contamination. Neglecting the precise information shared between these contexts could potentially lead to an amplification of erroneous information, will this have a negative impact on our task?

To address this concern, we introduce a novel parallel strategy that concurrently models both local and global contextual information while capturing their interaction. Specifically, we employ the existing Order-Aware (OA) block as the module for capturing local consensus and design a novel Bilateral Consensus Mining Attention (BCMA) block that operates in parallel with it, serving as the module for capturing global consensus. As demonstrated in Figure~\ref{fig:Introduce} (e), for each correspondence, the BCMA block not only establishes global dependencies but also highlights local information, akin to the matching process performed by the human central and peripheral vision. This embedding of local context is achieved by employing k-nearest neighbor (KNN) search separately in the query, key and value feature spaces to seek consistent neighbors. Subsequently, we concatenate the learned local consensus by OA block and global consensus by BCMA block in the channel dimension and facilitate their interaction through several additional ResNet blocks, resulting in a more dependable bilateral consensus. As shown in Figure~\ref{fig:Introduce} (d)-(f), through the acquisition of bilateral consensus, our network demonstrates remarkable performance in the correspondence pruning task.

In addition, we discover that inliers have higher consistency in the bilateral consensus feature space, a simple interaction operation is inadequate in the face of complex matching scenarios. To enhance the robustness of our network, we present the Bilateral Consensus Recalibrate (BCR) block to revalidate bilateral consensus at both local and global scales. To elaborate further, BCR block entails direct compression of bilateral consensus feature maps to generate a global scalar, along with the utilization of a KNN search search again to acquire a local vector. The fusion of these two tensors facilitates a soft selection process on the bilateral consensus, thereby amplifying the feature map's representational capacity. Finally, with all the proposed blocks, we formulate a Bilateral Consensus Learning Network tailored for the task of two-view correspondence pruning.

We summarize our contributions as follows:
\begin{itemize}
\item We propose a novel consensus learning strategy for the two-view correspondence pruning task. In contrary to previous progressive learning strategy, we concurrently learn local and global consensuses in parallel and obtain bilateral consensus by establishing interdependencies between them. To our knowledge, it is the first time leveraging bilateral consensus to handle the task of two-view correspondence pruning.
\item We propose a simple yet effective BCMA block as the global consensus learning module in bilateral consensus and a BCR block to rectify bilateral consensus. Through the process of learning and recalibration, our network is equipped to handle intricate matching scenarios.
\item We develop an effective BCLNet for correspondence pruning task. Extensive experiments demonstrate the effectiveness of our proposed BCLNet on the correspondence classification task and the camera pose estimation task. Noteworthily, BCLNet obtains 3.98\% mAP$5^{\circ}$ gains over the second best method on unknown outdoor dataset, and obviously accelerates model training speed.
\end{itemize}

\section{Related Work}
\subsection{Correspondence Pruning}
Over the years, correspondence pruning method has undergone extensive development and gradually divided into two branches, i.e., traditional hand-craft methods and learning-based methods. In traditional methods, hypothesize-and-verify strategy is a prevalent paradigm, $e.g.$, RANSAC~\cite{Fischler1981}, PROSAC~\cite{Chum2005}, MAGSAC~\cite{Barath2019},  LO-RANSAC~\cite{Lo-RANSAC}, NG-RANSAC~\cite{NG-RANSAC}, $etc$. To be specific, RANSAC~\cite{Fischler1981} repeatedly selecting a random subset of the data to fit models until the number of inliers meets a specified threshold. PROSAC~\cite{Chum2005} exploits LMI constraints to accelerate the model optimization process based on RANSAC, while MAGSAC~\cite{Barath2019} proposes $\sigma$-consensus to eliminate the need of user-defined threshold. But these methods are mostly task-specific, and the computation time adds exponentially with outliers increases, which is not enough to deal with the complex reality scenes.

Fortunately, the advent of deep learning makes it possible. As a pioneering work, LFGC~\cite{Yi2018} first proposes a PointNet-like architecture that treats correspondence pruning as a collection of correspondence classification and camera pose regression tasks. The impressive efficacy of it makes the adoption of permutation invariant pattern to handle correspondences a de facto standard. Subsequent works primarily focus on designing additional modules to capture richer context. Notably, OANet~\cite{Zhang2019} implicitly captures local context by mapping correspondences to clusters through differential pooling operation. CLNet~\cite{Zhao2021} initially proposes a local-to-global learning strategy and incorporates pruning operations. However, they only consider the improvement of a certain context or separately utilize detached contexts, ignoring the potential interactions between them, which limits the improvement of network performance.

\begin{figure*}[t]
\centering

\centerline{\includegraphics[width=1 \textwidth ]{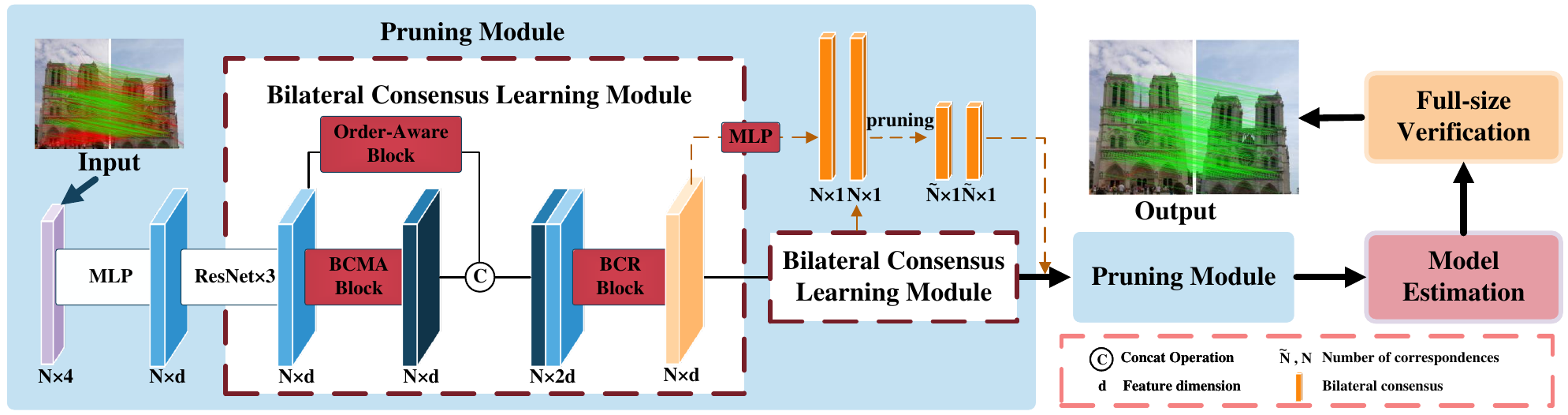}}
\caption{Architecture of BCLNet for correspondence pruning. We take putative correspondences $C \in  R^{N \times 4}$ as inputs and finally output the inlier probabilities $\omega \in  R^{N\times 1}$. The entire process involves two pruning modules, systematically refining correspondences into more reliable subsets. Each pruning module consists of our proposed Bilateral Consensus Mining Atttention block, Bilateral Consensus Recalibrate block and existing Order-Aware block.}
\label{fig:Framework}
\end{figure*}

\subsection{Consensus in correspondences}
The inliers between two images satisfy epipolar geometry constraint or homography transformation, while outliers are randomly distributed. This correspondence consensus has been studied predates the popularity of learning-based methods, $e.g.$, GMS~\cite{Bian2017} represents consensus as the statistical number of supported correspondences in a small area. LPM~\cite{Ma2019} considers the topological structure within the neighborhood of keypoints is always consistent. Inspired by these methods, learning-based methods also introduce the concept of consensus. CLNet~\cite{Zhao2021}, for instance, firstly differentiates between global and local consensus and proposes a local-to-global learning strategy. As an improved version of CLNet~\cite{Zhao2021}, NCMNet~\cite{Liu2023a} achieves significant improvements in exploring neighbor consistency in different feature spaces. ConvMatch~\cite{Zhang2023} maps correspondences onto grids and learns motion consistency using convolutions. In this paper, we also leverage the concept of consensus, but emphasize on interacting with different consensus information.

\section{Method}

\subsection{Problem Formulation}
Given a pair of images ($I$, $I'$), we first extract keypoints by existing feature detectors~\cite{Lowe2004,DeTone2018}. Then a nearest neighbor matching algorithm is used to obtain the initial correspondence set $C$:
\begin{equation}
C = \left\{c_{1}, c_{2}, c_{3}, \dots , c_{n}\right\} \in  R^{N\times 4} , c_{i} = (x_{i}, y_{i}, x_{i}', y_{i}')
\end{equation}
where $c_{i}$ is the i-th correspondence. $(x_{i}, y_{i})$ and $(x_{i}', y_{i}')$ represent the coordinates of keypoints that are normalized by camera intrinsics in the image $I$ and $I'$, respectively.
\par It is worth noting that the initial correspondence set $C$ is dominated by outliers. With $C$ as input, the goal of correspondence pruning is to predict the probability of each correspondence as an inlier and recover the relative camera motion. Specifically, reference to literature~\cite{Zhao2021}, BCLNet contains two pruning modules to progressive pruning $C$ into a more reliable candidate subsets $\hat{C}$. As show in Figure ~\ref{fig:Framework}, the complete architecture can be expressed as:
\begin{equation}
(\omega_{1} ,C_{1}) = {f}'_{\theta } (C) ,\quad (\omega_{2} ,{C}_{2}) = {f}'_{\varphi} (C_{1}),
\end{equation}
\begin{equation}
\hat{E} = g({C}_{2}, \omega_{2}), \quad \omega = h(\hat{E},C)
\end{equation}
where ${f}'_{\theta }(\cdot),{f}'_{\varphi}(\cdot)$ represent different network weights of the two pruning modules; $g(\cdot)$ denotes the weight eight-point algorithm; $h(\cdot)$ is full-size verification operation. Particularly, here $C_{1} \in  R^{N_{1}\times 4}$ and $C_{2} \in  R^{N_{2}\times 4}$ are pruned correspondence sets, in which $ N>N_{1}>N_{2}$; $\omega_{1} \in  R^{N\times 2}$ and $\omega_{2} \in  R^{N_{1}\times 2}$ are the weights predicted by bilateral consensus learning modules; $\omega \in  R^{N\times 1}$ is the inlier probabilities of the final prediction.

\subsection{Bilateral Consensus Mining Attention Block}
To achieve bilateral consensus, it is imperative to model local and global consensuses primarily. The existing Order-Aware block has demonstrated commendable accuracy in extracting local consensus. Consequently, an additional module must be fashioned to capture global consensus information. The potency of the self-attention mechanism in establishing global dependencies instills us with optimism. Therefore, we design a distinctive and lightweight Bilateral Consensus Mining Attention (BCMA) block. What's more, acknowledging the significance of local information in the correspondence pruning task, BCMA block embeds local context before acquiring global consensus.

\subsubsection{Local Context Embedding}
In contrast to conventional self-attention~\cite{Vaswani2017}, BCMA computes similarity score across channels, rather than in the spatial dimension, which gives it linear complexity while encodes the global context implicitly. We also introduce annular convolution~\cite{Zhao2021} to highlight local context before obtaining global attention map. From input tensor $X\in R^{N \times d}$, where $d$ represents channel dimensions, we first generate query $Q$, key $K$ and value $V$ via a layer normalization followed by MLP layers. Then, three neighbor embedding graphs are constructed using KNN search for each correspondence on value, query and key projections:
\begin{equation}
\mathcal{G}^{Q}_{i} = \left \{ \mathcal{V}^{Q}_{i}, \mathcal{E}^{Q}_{i}  \right  \}, \mathcal{G}^{K}_{i} = \left \{ \mathcal{V}^{K}_{i}, \mathcal{E}^{K}_{i}  \right  \}, \mathcal{G}^{V}_{i} = \left \{ \mathcal{V}^{V}_{i}, \mathcal{E}^{V}_{i}  \right  \}
\end{equation}

Towards graph $\mathcal{G}^{Q}_{i}$, nodes $\mathcal{V}^{Q}_{i} = \left \{c^{Q}_{i,1}, c^{Q}_{i,2}, \cdots, c^{Q}_{i,k} \right \}$, $c^{Q}_{i,m}$ denotes the m-th neighbor of $c^{Q}_{i}$ according to Euclidean distance between $c^{Q}_{i}$ and $c^{Q}_{m}$ in the query feature space. Edges $\mathcal{E}^{Q}_{i} = \left \{e^{Q}_{i,1}, e^{Q}_{i,2},\cdots, e^{Q}_{i,k} \right \} $ is formulated as~\cite{Wang2019}:
\begin{equation}
e^{Q}_{i,j} = \left [ z^{Q}_i, z^{Q}_i - z^{Q}_j \right ], \quad 1\le j\le k
\end{equation}
where $\left [ \cdot \right ] $  denotes the concatenation operation, $z^{Q}_i - z^{Q}_j$ represents residual feature between $c^{Q}_{i}$ and $c^{Q}_{j}$.

Whereafter, in order to effectively tapping local consensus through graph $\mathcal{G}^{Q} \in R^{N \times 2d} $, we categorize the neighbors into $g$ groups based on their affinity to the anchor node $c^{Q}_{i}$, where each group contains $k/g$ nodes. The final feature of $c^{Q}_{i}$ is then obtained by a convolution layer with $1 \times k/g$ kernels followed by a convolution layer with $1 \times g$ kernels. By aggregating all correspondences, we acquire a new feature mapping $Q_{local}$, that encapsulates local information. Let us do the same operation for $\mathcal{G}^{K}$ and $\mathcal{G}^{V}$, so that correspondence in each feature space aggregates different local contexts and we get $K_{local}, V_{local}$.

\subsubsection{Attention mechanism}
We alter the computation order of $Q, K, V$ in conventional self-attention ~\cite{Vaswani2017}, so that the similarity map A generated by the dot product of Q and K has only the size of $R^{d \times d}$, instead $R^{N \times N}$. Thus, its computational complexity is linear for the number of correspondences. The attention process is defined as:
\begin{equation}
\hat{X}  = W_{P} \cdot V_{local}\cdot Softmax\left ( Q_{local} \cdot K_{local}^{T} \cdot \alpha  \right ) +  X
\end{equation}
where $X$ and $\hat{X}$ are the input and output tensor; $W_{P}$ represents a MLP layer; $\alpha$ is a learnable scaling parameter to adjust the magnitude of the similarity map A.

After completing the above operations, the resulting feature $\hat{X}$ undergoes further refinement to achieve the desired global consensus, which involves an additional layer normalization and a ResNet block. Finally, we concatenate the obtained global consensus with the local consensus learned by the Order-Aware block in the feature dimension. This combined feature undergoes thorough interaction and recalibration through followed Bilateral Consensus Recalibrate block, resulting in the reliable bilateral consensus.

\subsection{Bilateral Consensus Recalibrate Block}
We find that inliers have higher consistency in the feature space after the convergence of bilateral consensus. However, when facing some challenging scenarios, the feature representation simply learned through a interaction operation is deficient. It remains susceptible to inaccuracies in the features when performing correspondence pruning. Hence, we propose to model the interdependence between channels as SENet~\cite{Hu2018} does to adaptively recalibrate the feature response across channels.

\begin{figure}[h]
\centering
\centerline{\includegraphics[width=0.5\textwidth]{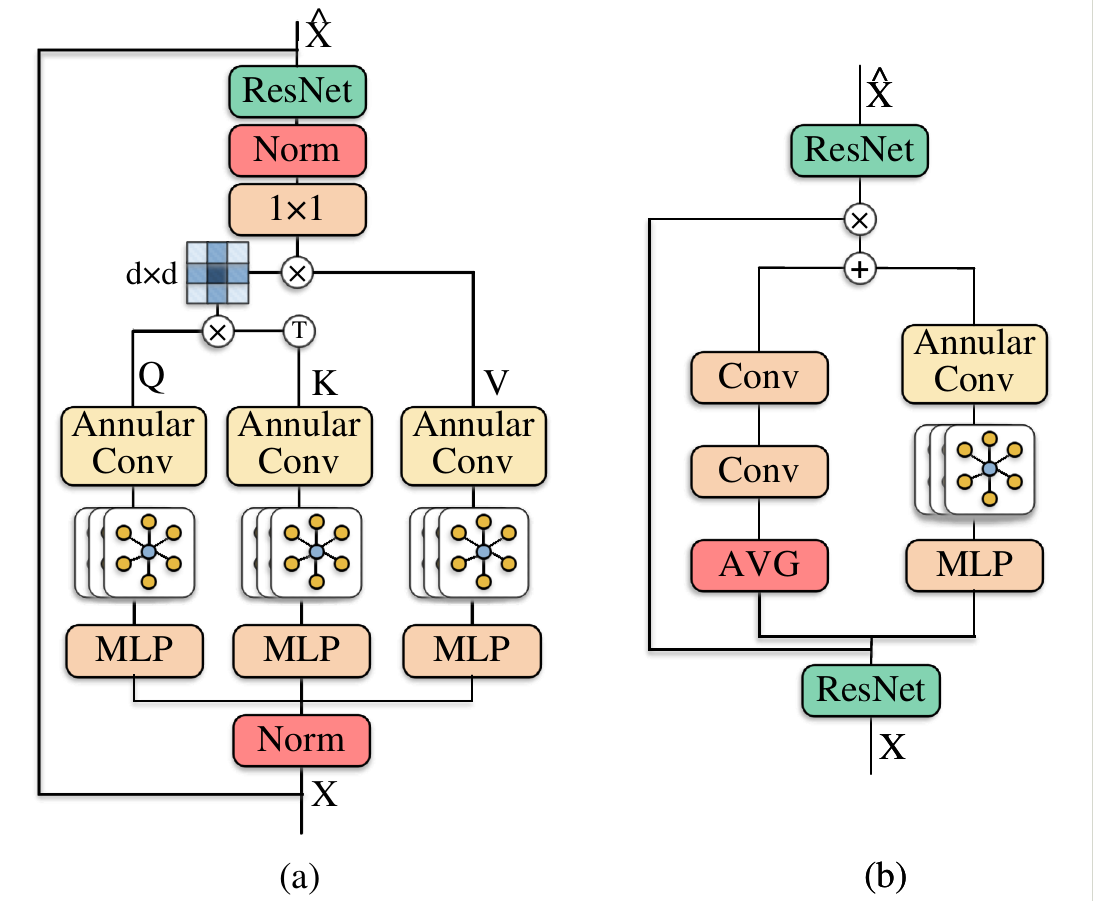}}
\caption{Illustration of the proposed (a) Bilateral Consensus Mining Attention (BCMA) block and (b) Bilateral Consensus Recalibrate (BCR) block.}
\label{fig:detail}
\end{figure}

For each feature map, SENet~\cite{Hu2018} compresses all spatial information into a scalar, which unfortunately neglects the vital local context containing motion consistency that is crucial for our task. To overcome this limitation, literature~\cite{Zheng2022} introduces a multi-scale attention block, but it only addresses this issue by using extra MLP layers to process correspondences individually. Inspired by this concept, we seize the opportunity to devise a Bilateral Consensus Recalibrate (BCR) block capable of capturing both local and global contexts concurrently.

Specifically, given the input feature $X \in R^{N \times d}$ , we first apply a global average pooling operation to obtain the global feature map $ {X}_{avg} \in R^{1 \times d}$. Thus, point-wise convolutional layers with a bottleneck structure are employed to get the global feature scalar $ {X}_{global} \in R^{1 \times d}$. The operations are formulated as:
\begin{equation}
{X}_{global} = Conv_1\left ( Conv_2\left ( AVG\left (  X\right )  \right )  \right )
\end{equation}
where the weights of $Conv_1$ and $Conv_2$ are $d \times d/r \times 1 $ and $d/r \times d \times 1 $ respectively; $r$ is the channel reduction ratio; ${X}_{global}$ is the recalibrated global scalar. For simplicity, we omit the Batch Normalization layer and ReLU layer.

Next, considering the strong consistency of inliers at the bilateral consensus level, we adopt the approach that mentioned in the previous local context embedding section to get the local feature vector ${X}_{local}$. At last, we combine the features ${X}_{global}$ and ${X}_{local}$ and apply a sigmoid function to obtain the final weight for softly selecting $X$ by an element-wise multiplication operator. The entire process is represented as:
\begin{equation}
{X}_{out} = Sigmod\left ( {X}_{local}+{X}_{global} \right )\cdot X
\end{equation}
What's more, to effectively manage the bilateral consensus features post recalibration, we also add three Resnet blocks at the end to adequately handle the features.

\subsection{Network Framework}
The overall architecture of the BCLNet proposed by us is shown in Figure~\ref{fig:Framework}. BCLNet primarily consists of two pruning modules, a camera motion estimation module, and a full-size verification module. The camera motion estimation module is responsible for computing essential matrix from the last pruned reliable correspondence subset and subsequently passes it to the full-size verification module for classification of all correspondences.

Principally, each pruning module is composed of a MLP layer, three ResNet blocks and two Bilateral Consensus Learning Modules. We use MLP layer to elevate features to higher dimensional spaces d and ResNet block serving as the basic module for extracting correspondence features, which incorporates several MLP layers and normalization layers. An Order-Aware block is incorporated into the Bilateral Consensus Learning module alongside the proposed BCMA block, allowing simultaneous extraction of both local and global contextual information. Subsequently, these two forms of contextual information are combined to form bilateral consensus, which is then revalidated using the BCR block. Combining the above modules, we get a progressive correspondence pruning network.

\subsection{Loss Function}
Follow~\cite{Yi2018}, to balance the correspondence classification task and the camera pose estimation task, we develop a hybrid loss function to optimize the proposed BCLNet:
\begin{equation}
\mathcal{L}=\mathcal{L}_{cls} + \lambda \mathcal{L}_{ess}
\label{loss}
\end{equation}
where $\mathcal{L}_{cls}$ represents correspondence classification loss, $\mathcal{L}_{ess}$ denotes essential matrix estimation loss, $\lambda$ is the weight factor used to balance the two losses.
\par In $\mathcal{L}_{cls}$, to mitigate the effects of label ambiguity, we add adjustable temperature $\tau$~\cite{Zhao2021}. For inliers, $\tau_{i}$ is negatively correlated with the epipolar distance $d_{i}$, while for outliers $\tau_{i} = 1$. The loss $\mathcal{L}_{cls}$ is given as:
\begin{equation}
\mathcal{L}_{cls}(\omega, y)=\sum\limits_{i=1}^{N_k}\mathcal{H}( \tau_{i}\odot{\omega}_{i},y_{i}) + \mathcal{H}( \hat{\tau} \odot\hat{\omega},\hat{y})
\end{equation}
where ${\omega}_{i}$ is the predicted logit vector of the $i-th$ bilateral consensus learning module; $\hat{\omega}$ indicates the output of the last MLP layer; $y_{i}$ and $\hat{y}$ are ground-truth labels; $\mathcal{H}$ indicates the binary cross entropy loss function; $\odot$ denotes the Hadamard product and ${N_k}$ represents the num of bilateral consensus learning modules.

$\mathcal{L}_{ess}$ can be write as a geometry loss~\cite{Ranftl2018}:
\begin{tiny}
\begin{equation}
\mathcal{{L}}_{ess}(\hat{{E}}, {E}) = \frac{({p{\prime}}^T\hat{E}{p})^2}{\parallel E{p}\parallel _{[1]}^2 + \parallel{E}{p}\parallel _{[2]}^2  + \parallel{E}^{T}{p{\prime}}\parallel  _{[1]}^2 + \parallel {{E}}^{T}{p{\prime}}\parallel_{[2]}^2 }
\end{equation}
\end{tiny}where $\hat{E}$ and $E$ denote the predicted essential matrix and ground-truth essential matrix respectively; $\parallel\cdot\parallel_{[i]}$ represents the i-th element of this vector; $p$ and $p{\prime}$ are the virtual correspondence coordinates generated by using $E$.

\section{Experiments}
We first introduce the datasets and evaluation metrics used in the experiment. Subsequently, by comparing with some state-of-the-art methods, we validate the effectiveness of BCLNet in both correspondence classification task and camera pose estimation task. At last, we conduct some ablation experiments to discuss the setting of hyperparameters and evaluate the role of each module.

\subsection{Evaluation Protocols}
\subsubsection{Dataset}
Towards an effort to demonstrate the effectiveness of BCLNet, we conduct experiments on both indoor and outdoor scenes. For the outdoor dataset, we utilized Yahoo's YFCC100M~\cite{Thomee2016}, a vast collection containing 100 million pieces of multimedia data. As for the indoor setting, we relied on SUN3D~\cite{Xiao2013}, which is an RGBD video dataset encompassing entire rooms. Followed the data division approach outlined in ~\cite{Zhang2019}, we train all models at the same training setting to ensure an equitable comparison.

\subsubsection{Evaluation Metrics}
Our proposed network addresses two key tasks: correspondence classification and camera pose estimation. For classification task, we employ precision (P), recall (R), and F-score (F) as evaluation metrics to measure the performance of classification task. Precision is the ratio of correctly identified correspondences to preserved correspondences, while Recall is the ratio of correctly identified correspondences to all correct correspondences in the initial set. The F-score combines Precision and Recall for a comprehensive measure, defined as $2 * P * R / (P + R)$. For camera pose estimation task, we use mean average precision (mAP) of angular differences under various error thresholds as evaluation metric. The angular difference is determined by comparing the rotation and translation vectors predicted by our model with the corresponding ground truth values.

\begin{table}[t]
 \centering
 \scalebox{0.8}{
  \begin{tabular}{c c c c c c c }
   \toprule
   \multirow{2}*{Methods} & \multicolumn{3}{c}{Known Scene} & \multicolumn{3}{c}{Unknown Scene} \\
   \cmidrule(r){2-4} \cmidrule(r){5-7}
   & $P$ (\%) & $R$ (\%) & $F$ (\%) & $P$ (\%) & $R$ (\%) & $F$ (\%)    \\
   \midrule
   RANSAC           & 47.44 & 52.64 & 49.90 & 43.51 & 50.68 & 46.82         \\
   PointNet++       & 49.84 & 86.41 & 63.22 & 46.60 & 84.17 & 59.99          \\
   LFGC             & 56.64 & 86.30 & 68.39 & 54.67 & 84.76 & 66.47          \\
   OANet++          & 60.03 & 89.31 & 71.80 & 55.78 & 85.93 & 67.65         \\
   MSA-Net          & 61.98 & 90.53 & 73.58 & 58.70 & 87.99 & 70.42         \\
   CLNet            & 76.04 & 79.27 & 77.62 & 75.05 & 76.41 & 75.72         \\
   MS$^2$DG-Net     & 63.17 & 90.98 & 74.57 & 59.11 & 88.40 & 70.85         \\
   ConvMatch        & 63.07 & \textbf{91.55} & 74.69 & 58.77 & \textbf{89.39} & 70.92       \\
   NCMNet           & \textbf{78.49} & 81.72 & 79.69 & 77.07 & 78.27 & 77.41     \\
   Ours             & \textbf{78.49} & 82.56 & \textbf{80.10} &\textbf{77.39}  & 79.77 & \textbf{78.31}   \\
   \bottomrule
 \end{tabular}}
 \caption{Performance comparisons of our network and other models about the Precision, Recall and F-score on the YFCC$100$M dataset in the correspondence classification task.}
 \label{tab:prfcompare}
\end{table}
\subsection{Implementation Details}
In our experiments, feature detectors(SIFT~\cite{Lowe2004}, Superpoint~\cite{DeTone2018}) and nearest neighbor matching method are adopted to establish $N = 2000$ initial correspondences and then our network. We go through two pruning modules with a pruning ratio of 0.5 and end up with a reliable set of correspondences of $N/4$. Throughout both pruning modules, we increase the feature dimension $d$ to 128. For the first pruning module, we use the initial set of correspondences as input and the number of $k$ neighbors is empirically set to 9. For the second pruning module, we set $k$ to 6 and use the weights predicted in the previous module along with the pruned correspondences set as input. Within the BCMA layer and BCR layer of the two pruning modules, we set the number of groups $g$ to 3 and 2, respectively. Additionally, in the Order-Aware block, we set the cluster number to 150 for outdoor scenes and 250 for indoor scenes. All networks are implemented in PyTorch~\cite{Paszke2019} and trained using the Adam optimizer~\cite{Kingma2014} with an initial learning rate of $10^{-3}$ and a batch size of 32. The training process consists of 500k iterations. In Equation \ref{loss}, the weight $\lambda$ is initialized to 0, and then fixed at 0.5 after the first 20k iterations.

\subsection{Correspondence Classification}
We compare BCLNet with RANSAC~\cite{Fischler1981} and recent state-of-the-art learning-based methods for both known and unknown outdoor scenes, which included PointNet++~\cite{Qi2017}, LFGC~\cite{Yi2018}, OANet~\cite{Zhang2019}, MSANet~\cite{Zheng2022}, CLNet~\cite{Zhao2021}, MS2DNet~\cite{Dai2022}, ConvMatch~\cite{Zhang2023}, and NCMNet~\cite{Liu2023a}. Table~\ref{tab:prfcompare} presents the comparative results in the classification task, measured using Precision, Recall, and F-score metrics. For all correspondences, we consider them as inliers if their epipolar distance is less than a certain threshold (10$^{-4}$).

\begin{table}[t]
\centering
\scalebox{0.8}{
\begin{tabular}{lcccc}

\toprule	
\multirow{2}{*}{Methods}   & \multicolumn{2}{c}{YFCC100M($\%$)}  & \multicolumn{2}{c}{SUN3D($\%$)} \\
\cmidrule(r){2-3} \cmidrule(r){4-5} & Known & Unknown & Known & Unknown  \\
\midrule
         RANSAC            & 5.81     & 9.07     & 4.52     & 2.84      \\
         PointNet++        & 10.49    & 16.48    & 10.58    & 8.10      \\
         LFGC              & 17.45    & 25.95    & 11.55    & 9.30      \\
         OA-Net++          & 32.57    & 38.95    & 20.86    & 16.18     \\
         MSA-Net           & 39.53    & 50.65    & 18.64    & 16.86     \\
         CLNet             & 38.27    & 51.80    & 19.20    & 15.83     \\
         MS$^2$DG-Net      & 38.36    & 49.13    & 22.20    & 17.84     \\
         ConvMatch         & 43.48    & 54.62    & \textbf{25.36}    & \textbf{21.71}     \\
         NCMNet            & 49.10    & 62.10    & 24.91        & 19.57     \\
         \textbf{BCLNet(Ours)}   & \textbf{52.62}   & \textbf{66.08}     & 24.59     & 19.96   \\
\bottomrule	
\end{tabular}
}
\caption{Comparative results about our network and other networks when using SIFT as feature extractor on the YFCC$100$M and SUN3D datasets, with mAP5$^\circ$ ($\%$) is reported.}
\label{tab:different scenes}
\end{table}

\begin{figure}[t]
	\centering
    \subfigure[RANSAC]{
    \begin{minipage}[b]{0.32\linewidth}
    \includegraphics[width=1\linewidth]{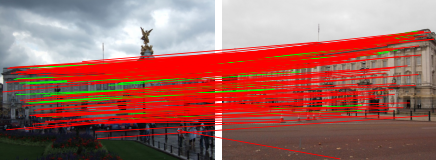}
    \includegraphics[width=1\linewidth]{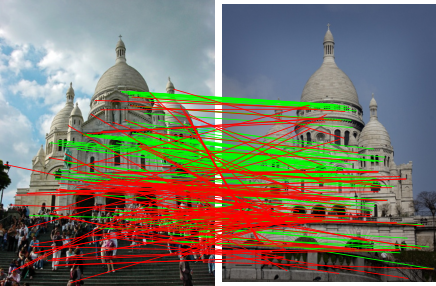}
    \includegraphics[width=1\linewidth]{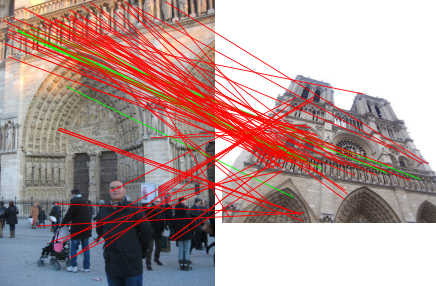}
    \includegraphics[width=1\linewidth]{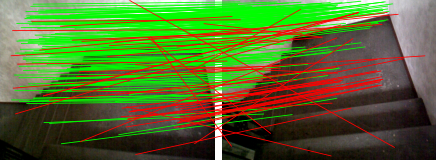}
    \includegraphics[width=1\linewidth]{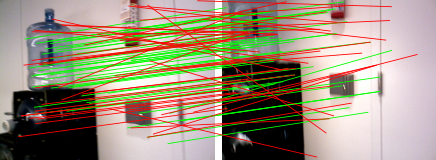}
    \end{minipage}    \hspace{-2mm}}
    \subfigure[CLNet]{
    \begin{minipage}[b]{0.32\linewidth}
    \includegraphics[width=1\linewidth]{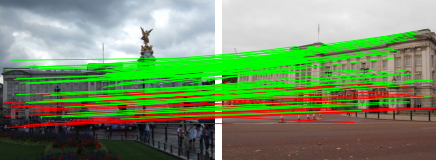}
    \includegraphics[width=1\linewidth]{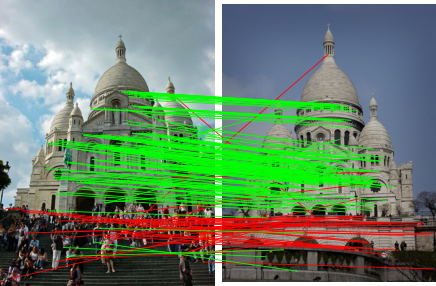}
    \includegraphics[width=1\linewidth]{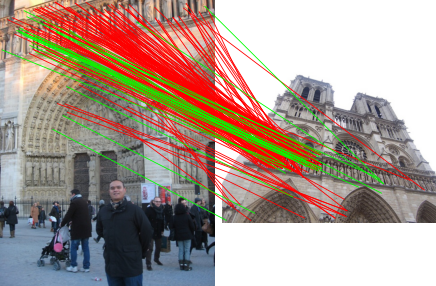}
    \includegraphics[width=1\linewidth]{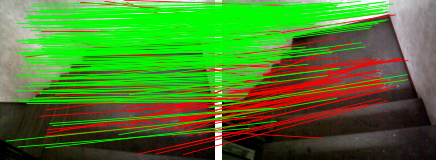}
    \includegraphics[width=1\linewidth]{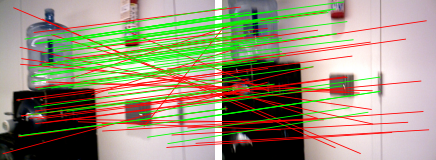}
    \end{minipage}   \hspace{-2mm}}
    \subfigure[BCLNet(Ours)]{
    \begin{minipage}[b]{0.32\linewidth}
    \includegraphics[width=1\linewidth]{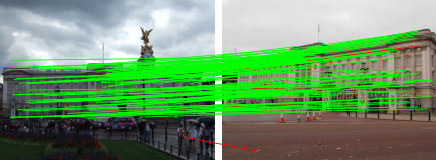}
    \includegraphics[width=1\linewidth]{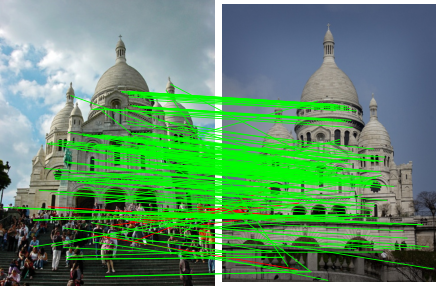}
    \includegraphics[width=1\linewidth]{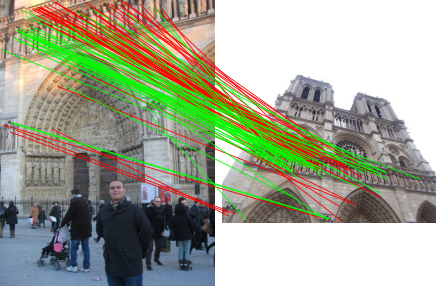}
    \includegraphics[width=1\linewidth]{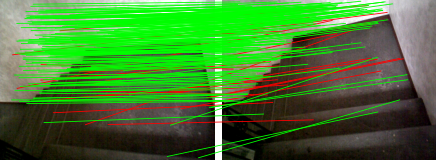}
    \includegraphics[width=1\linewidth]{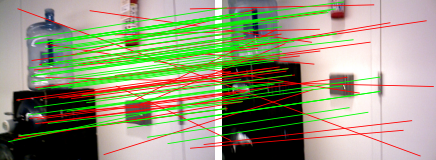}
    \end{minipage}}

\caption{Visualization results of two-view correspondence pruning on the unknown outdoor scenes and unknown indoor scenes. From left to right are the results of RANSAC, CLNet and BCLNet, respectively. Inliers(green lines) and outliers(red lines) retained by algorithms are exhibited. }
\label{fig:visiual results}
\end{figure}

From Table~\ref{tab:prfcompare}, we can observe that BCLNet achieves the best results on Precision and F-score metrics across all scenarios. However, it is worth noting that both our method and CLNet~\cite{Zhao2021}, NCMNet~\cite{Liu2023a} show a significant decrease on Recall compared to other learning-based networks. This reduction is attributed to the adoption of pruning strategy, leading to the removal of inliers during the pruning process. Although full-size verification restores some inliers, the predicted geometric model still implicitly inherits this attribute. Specifically, NCMNet~\cite{Liu2023a} represents an improved version of CLNet~\cite{Zhao2021}, capturing richer local consensus and being the most effective model prior to our work. But our network demonstrates improvements over NCMNet~\cite{Liu2023a} in all performance indicators, proving that learning bilateral consensus effectively suppresses the spread of erroneous context. Furthermore, with comparable parameters, the training time of BCLNet has been reduced by nearly 25 hours compared to NCMNet~\cite{Liu2023a}, thanks to our key improvements in the attention mechanism. Additionally, we also visualized some visualization results of RANSAC, CLNet and BCLNet in Figure~\ref{fig:visiual results}. As can be seen, BCLNet achieves the best results.

\subsection{Camera Pose Estimation}
The camera pose estimation aims to calculate the positional changes (translation and rotation) between two cameras when capturing an image pair. We compare BCLNet with other methods described in the previous section on both indoor and outdoor scenes to verify the generalization of our network. In addition, since the quality of the initial correspondences will inevitably affect the effect of camera pose estimation, we also utilize traditional (SIFT) and learn-based methods (SuperPoint) as feature extractors respectively to prove that the network has strong robustness.

As shown in Table~\ref{tab:different scenes}, the comparison results for indoor and outdoor scenes are presented. We report mAP5$^\circ$ (\%) for both known and unknown scenes. It is evident that BCLNet achieves state-of-the-art performance in nearly all scenarios, particularly with significant improvements of $3.52\%$ and $3.98\%$ in outdoor known and unknown scenarios compared to the second best method respectively, while maintaining lower computational complexity. In indoor scenarios, our performance slightly lags behind that of ConvMatch, albeit with the advantage of our model size being almost one-third smaller.

In addition, Table~\ref{tab:feature extraction method} presents comparison results obtained using learning-based feature extractor and hand-craft feature extractor on the YFCC100M dataset, with mAP5$^\circ$ ($\%$) and mAP20$^\circ$ ($\%$) reported. BCLNet continues to demonstrate superior performance across all scenarios, affirming the robustness of our network to the quality of the initial correspondence set.

\begin{table}[t]
\centering
\scalebox{0.8}{
\begin{tabular}{clcccc}
	\toprule
\centering
&\multirow{2}{*}{Matcher} & \multicolumn{2}{c}{Known Scene} &\multicolumn{2}{c}{Unknown Scene} \\
\cmidrule(r){3-4} \cmidrule(r){5-6} & &5$ ^\circ$($\%$) &20$ ^\circ$($\%$) &5$ ^\circ$($\%$) &20$ ^\circ$($\%$)\\
\midrule
\multirow{10}{*}{SIFT}
        & RANSAC            & 5.81     & 16.88    & 9.07     & 22.92     \\
        & PointNet++        & 10.49    & 31.17    & 16.48    & 42.09     \\
        & LFGC              & 13.81    & 35.20    & 23.95    & 52.44     \\
        & OA-Net++          & 32.57    & 56.89    & 38.95    & 66.85     \\
        & MSA-Net           & 39.53    & 61.75    & 50.65    & 77.99     \\
        & CLNet             & 38.27    & 62.48    & 51.80    & 75.76     \\
        & MS$^2$DG-Net      & 38.36    & 64.04    & 49.13    & 76.04     \\
        & ConvMatch         & 43.48    & 66.14    & 54.62    & 77.24     \\
        & NCMNet            & 49.10    & 70.80    & 62.10    & 81.67     \\
        & \textbf{BCLNet(Ours)}     & $\textbf{52.62}$   & $\textbf{72.89}$     & $\textbf{66.08}$     & $\textbf{83.38}$     \\
\midrule
\multirow{10}{*}{SuperPoint}
        & RANSAC            & 12.85    & 31.22     & 17.47    & 38.83     \\
        & PointNet++        & 11.87    & 33.35     & 17.95    & 49.32     \\
        & LFGC              & 12.18    & 34.75     & 24.25    & 52.70     \\
        & OA-Net++          & 29.52    & 53.76     & 35.27    & 66.81     \\
        & MSA-Net           & 30.63    & 53.74     & 38.53    & 68.56     \\
        & CLNet             & 27.56    & 50.82     & 39.19    & 67.37     \\
        & MS$^2$DG-Net      & 31.15    & 55.16     & 39.19    & 70.36     \\
        & ConvMatch         & 38.34    & 60.25     & \textbf{48.80}    & 74.59     \\
        & NCMNet            & 40.33    & 62.37     & 48.60    & 75.38     \\
        & \textbf{BCLNet(Ours)}    & $\textbf{40.56}$   & $\textbf{62.71}$     & 48.07     & $\textbf{75.84}$     \\
\bottomrule
\end{tabular}
}
\caption{Comparative results about our network and other networks when using SIFT and SuperPoint as feature extraction methods on the YFCC$100$M dataset. mAP5$^\circ$ ($\%$) and mAP20$^\circ$ ($\%$) are reported.}
\label{tab:feature extraction method}
\end{table}

\subsection{Ablation}
In this section, we conduct ablation studies on the unknown scene of YFCC100M to verify the role of each key component in our network. We also perform some parameter analysis to balance performance and efficiency.

\subsubsection{Main components}
The proposed BCLNet comprises three main modules $i.e.$ BCMA block, BCR block, and OA block. Utilizing the pruning strategy from the reference ~\cite{Zhao2021} as a baseline, we evaluate the performance gains of these key components. The BCMA block is employed to run in parallel with the OA block to separately obtain local and global consensus, while the BCR block is used to re-verify bilateral consensus. As depicted in Table ~\ref{tab:main compnent}, with the integration of each module, the network's performance shows gradual improvement. When all modules are combined, we achieve the best result.

\begin{table}[t]
\centering
\scalebox{0.8}{\begin{tabular}{cccc|cc}

\toprule
     IPS        & BCMA      & BCR       & OA        & MAP5$^\circ$($\%$)    & MAP20$^\circ$($\%$)   \\
\midrule
    $\surd$     &           &           &           & 51.80           & 75.76           \\
    $\surd$     & $\surd$   &           &           & 59.98           & 80.72           \\
    $\surd$     & $\surd$   & $\surd$   &           & 61.03           & 81.18           \\
    $\surd$     & $\surd$   & $\surd$   & $\surd$   & 66.08           & 83.38           \\
\bottomrule

\end{tabular}}
\caption{Ablation studies regarding performance gains of the key components in each pruning module on outdoor dataset. mAP5$^\circ$ ($\%$) and mAP20$^\circ$ ($\%$) are reported. IPS: the iterative pruning strategy. BCMA: the Bilateral Consensus Mining Attention block. BCR: the Bilateral Consensus Recalibrate block. OA: the Order-Aware block.}
\label{tab:main compnent}
\end{table}

\begin{figure}[t]
\centering
\centerline{\includegraphics[width=0.27\textwidth ]{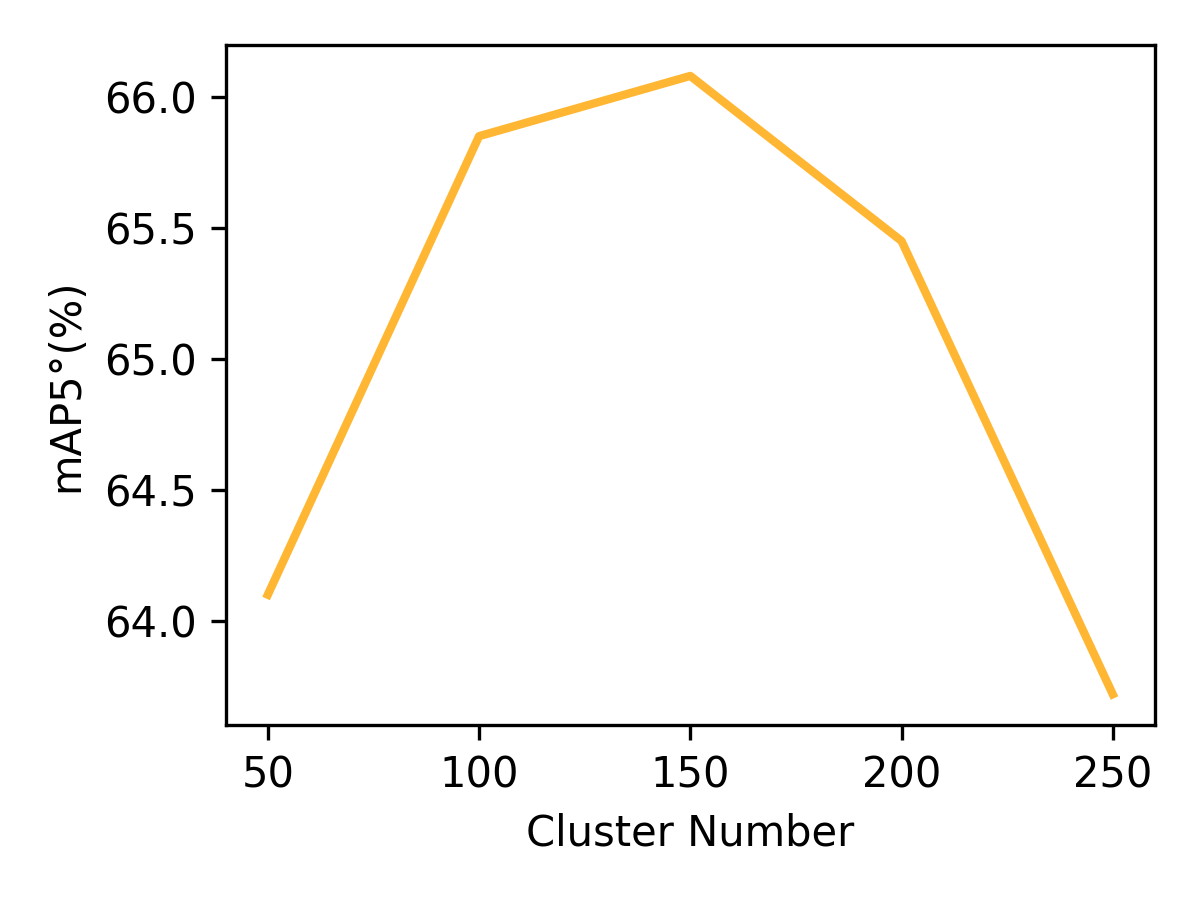}}
\caption{Impact of cluster number on BCLNet performance.}
\label{fig:Cluster}
\end{figure}

\subsubsection{Parameter Analysis}

In ours experiments, we observe a significant enhancement in the network's performance upon incorporating the Order-Aware block. This is attributed to the importance of local context in the correspondence pruning task. As bilateral consensus seems to weaken the local consensus, therefore, we adjust cluster number in the Order-Aware block to compensate for this deficiency. Intuitively, higher values of the cluster in the Order-Aware block are expected to improve network performance. However, this also bring in increased network parameters and computational burden. To strike a balance between model size and efficiency, we conduct tests to analyze the impact of different cluster values on outdoor dataset, as shown in Figure ~\ref{fig:Cluster}. Ultimately, we set it to 150 on outdoor scenes.

\section{Conclusion}
In this paper, we propose an effective Bilateral Consensus Learning Network (BCLNet) to cope with correspondence pruning problem. Considering the complementarity between local and global contexts, we interact these two types of information to get a bilateral consensus for accurately identify inliers. Simultaneously, to handle complex matching scenarios and mitigate the impact of incorrect information, we further design a Bilateral Consensus Recalibrate block that enhances the feature representation capability. Numerous experiments conducted on public benchmarks consistently demonstrate the superiority of our method over the current state-of-the-art technologies.

 \section*{Acknowledgment}%
{\small This work was supported by the National Natural Science Foundation of China under Grants 62072223, 62125201 and 62020106007.

\bibliography{BCLNetbib}}
\end{document}